\documentclass[runningheads]{llncs}
\usepackage[T1]{fontenc}
\usepackage{graphicx}
\usepackage{booktabs}
\usepackage[misc]{ifsym}
\newcommand{\corr}{(\Letter)}
\usepackage{mwe}
\usepackage{graphicx}
\usepackage{booktabs}
\usepackage{latexsym}
\usepackage{enumerate,enumitem}
\usepackage{comment}
\usepackage{xcolor}
\usepackage{fontawesome5}
\usepackage{tikz}
\usetikzlibrary{arrows.meta,positioning,fit,backgrounds}
\usepackage{url}
\usepackage{hyperref}
\usepackage{cleveref}

\begin{document}

\title{Virgil: Navigating Explainability for Transformer-based Language Models}

\titlerunning{Virgil: Navigating Explainability for Transformer-based Language Models}


\author{Martino Ciaperoni\inst{1} \corr \and
Sezer Kutluk\inst{1} \and
Benedetta Muscato\inst{1} \and
\\ Marta Marchiori Manerba\inst{2} \and
Fosca Giannotti\inst{1}
}


\authorrunning{M. Ciaperoni et al.}

\institute{Scuola Normale Superiore, Pisa, Italy \email{\{name.surname\}@sns.it}
\and
University of Turin, Italy \email{\{name.surname\}@unito.it}}

\toctitle{Virgil: Navigating Explainability for Transformer-based Language Models}
\tocauthor{Martino Ciaperoni, Sezer Kutluk, Benedetta Muscato, Marta Marchiori Manerba, Fosca Giannotti}

\maketitle              
\setcounter{footnote}{0}

\begin{abstract}
Explainability for transformer-based language models is becoming crucial as these systems are deployed in high-stakes applications. As a result, the ecosystem of explainability tools is rapidly evolving, becoming richer, but also more fragmented and harder to navigate. 

To address this challenge, we present Virgil, an interactive system that lets practitioners and researchers, including non-experts, navigate explainability tools for transformer language models. Supported by a curated knowledge base, the system enables users to discover and compare explainability tools within a unified interface.

\end{abstract}

\section{Introduction}
Transformer-based language models are widely deployed across many applications, including high-stakes domains such as law and healthcare~\cite{zhao_survey}. However, their decision processes remain difficult to interpret, raising concerns about hallucinations and unexpected behaviors~\cite{zhao_survey}. As a result, explainability for language models has attracted growing attention, leading to a rapidly expanding landscape of explainability tools (or \emph{explainers}), ranging from intuitive input attributions to mechanistic interpretability~\cite{ferrando2024primer}. 
Navigating this landscape can be challenging. Although surveys offer useful overviews and taxonomies, they rarely provide guidance for selecting suitable explainers in real-world settings~\cite{stakeholders_survey,zhao_survey}.

To bridge this gap, we introduce Virgil, an interactive system designed to guide users through the landscape of explainers for transformer-based language models. Named after the guide who leads Dante through the \emph{Inferno}, 
Virgil helps users identify explainers suited to their needs through structured filters or natural-language queries over a curated knowledge base of diverse explainers. Retrieved explainers are presented through concise descriptions that highlight their key characteristics and trade-offs.
In addition, Virgil enables users to directly run selected explainers and compare their outputs.
Thanks to its modular architecture, Virgil can be easily extended as new explainers emerge, positioning it as a central resource for explainability of language models across both academia and industry.
Virgil can be accessed at \url{https://huggingface.co/spaces/Explainability4LanguageModels/Virgil} and a video demonstration can be found at \url{https://www.youtube.com/watch?v=Ybs8IYztL8k}.

\section{Virgil Architecture}
Virgil is implemented in Python and provided as a web application built with \texttt{Streamlit}\footnote{\url{https://streamlit.io}}. The source code is publicly available at \url{https://github.com/maciap/Virgil}, which also contains the full list of explainers included in Virgil.

\begin{figure}[t]
    \centering
    \includegraphics[width=0.75\textwidth]{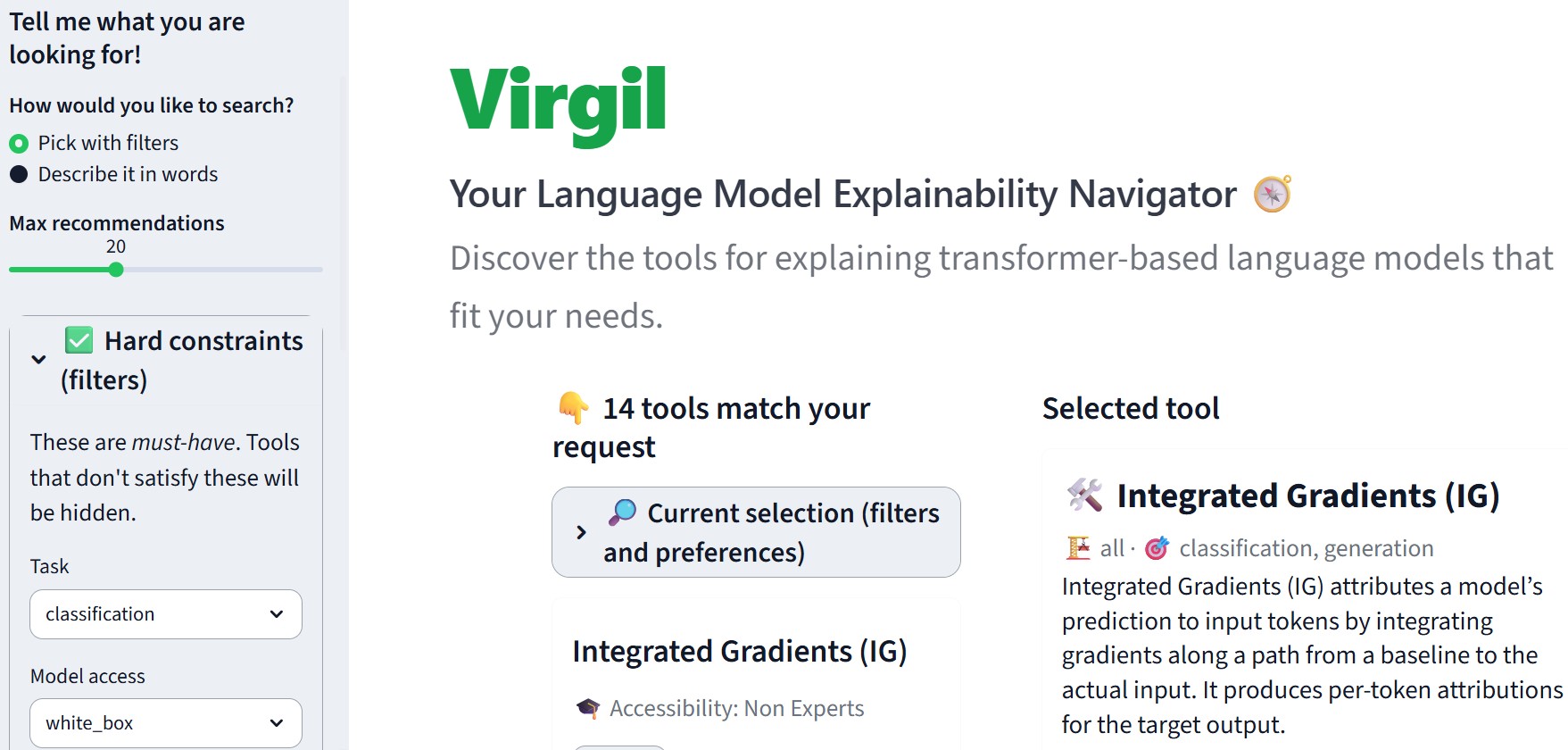}
    \vspace{1em}   
\includegraphics[width=0.8\textwidth]{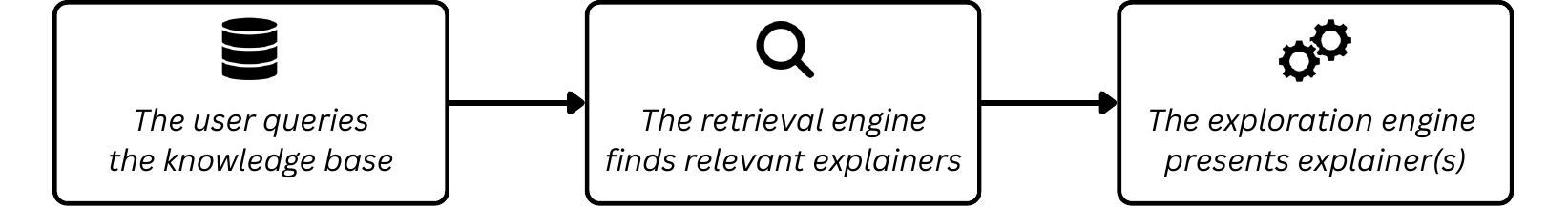}
    \caption{Overview of the Virgil interface. Users query the knowledge base on the left. Matching explainability tools are displayed in the center, while detailed information about selected explainers is shown on the right.
    }
    \label{fig:architecture}
\end{figure}

\noindent \textbf{Overview.} Virgil follows a modular architecture with three components: a \textit{knowledge base} storing structured descriptions of explainers, a \textit{retrieval engine} that identifies explainers matching user requirements, and an \textit{exploration engine}  that allows users to interactively explore the identified explainers.   
As illustrated in~\Cref{fig:architecture},
Virgil is accessible through a user-friendly interface.

\noindent \textbf{Knowledge base.} At the core of the system lies a curated knowledge base of explainers, each represented by an \emph{explainer card}. At the time of writing,
the knowledge base contains 43 explainer cards. 
Cards contain structured fields describing the target macro-task (text classification or generation), model access (white-box or black-box), supported transformer architectures (encoder-only, decoder-only, or encoder-decoder), explanation scope (local or global), and expertise required to understand the explanations (non-experts, mid-experts, or experts). For each explainer, they also contain an overview, capabilities, strengths, limitations, useful references and a link to an implementation.

\noindent \textbf{Retrieval engine.}  The retrieval engine maps user requirements to suitable explainers. Virgil supports structured queries (i.e., filters on task, model access, architecture, and explanation scope) and free-text queries describing user needs. Free-text queries are embedded using a sentence-transformer model (all-MiniLM-L6-v2~\cite{reimers2019sentence}) and compared with three textual fields of explainer cards, namely overview, capabilities, and strengths. Similarity scores are aggregated through a weighted ranking that assigns weights of 0.5, 0.4, and 0.1 to the three fields, respectively. Results can also be ranked by expertise requirements.

\noindent \textbf{Exploration engine.} 
After retrieving explainers matching user needs, users can select one to explore. The exploration engine presents their key characteristics and, when available, enables interactive execution of explainers with custom inputs, Hugging Face pretrained models~\cite{wolf2020transformers} and parameters, producing explanations and useful visualizations. 
It also supports side-by-side comparison of explainers (\emph{comparative view}).

\section{Using Virgil}
Virgil can be used online without coding or run locally with hardware acceleration support. 
Users interact with the system by querying the knowledge base to retrieve relevant explainers, which can then be selected, explored, and optionally executed.

\noindent \textbf{Target audience.}
Virgil supports a broad range of users interested in explaining transformer-based language models, from practitioners with limited experience in explainability to researchers conducting preliminary analyses and exploring or comparing different explainers. The system can also support educational, training, and collaborative activities involving multiple stakeholders.

\noindent \textbf{Preliminary user feedback.}
We gathered preliminary feedback from 10 anonymous researchers through a survey\footnote{https://forms.gle/nsME36PGQvum7Fh3A} assessing Virgil intuitiveness, comprehensibility, and usefulness. 
Respondents reported that Virgil is intuitive (80\% agree or strongly agree) and that the explainer descriptions are clear and useful (70\% agree or strongly agree). Most participants also indicated that they would recommend Virgil to colleagues (90\% likely or very likely). 
While encouraging, these results are limited by the small number of participants and their research-oriented background.
We plan to extend this evaluation to a larger group of practitioners and researchers in future work.

\begin{figure}[t!]
    \centering
    \begin{tabular}{c|c}
       \raisebox{-0.1em}{\includegraphics[width=0.38\linewidth]{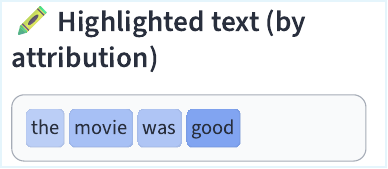}} 
       &  
       \includegraphics[width=0.41\linewidth]{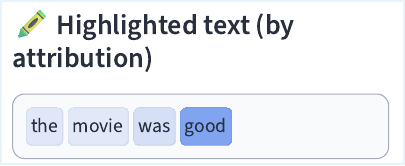} \\

       \textsc{Input×Gradient}
       & \textsc{Integrated Gradients}
    \end{tabular}
    \caption{Token-level attribution scores computed in Virgil using \textsc{Input×Gradient} (left) and \textsc{Integrated Gradients} for the  example input "The movie was good".}
    \label{fig:use_case_attribution}
\end{figure}

\noindent \textbf{Illustrative use case.}
As an example, consider a sentiment classification task on movie reviews where a user seeks to identify the most important tokens driving the prediction. The user queries Virgil for explainers compatible with text classification models, specifying an encoder-only transformer, white-box access, and local explanations. The system retrieves 16 results. The user first runs \textsc{Input×Gradient}~\cite{DBLP:journals/corr/SimonyanVZ13} using a pretrained sentiment classifier. However, the attribution scores appear unstable across similar inputs. The user therefore switches to the comparison view and evaluates \textsc{Integrated Gradients}~\cite{DBLP:conf/icml/SundararajanTY17}, which produces more stable attributions. \Cref{fig:use_case_attribution} shows \textsc{Integrated Gradients} more clearly highlighting ``good'' as driving the positive prediction. The user then queries ``counterfactual'' and runs \textsc{Polyjuice}~\cite{wu2021polyjuice} to generate a counterfactual example that flips the prediction. The analysis can be further extended by inspecting internal representations to study how positive sentiment is encoded. The same workflow applies to many applications, including in high-stakes domains.

\section{Conclusion}
We introduced Virgil, an interactive system for discovering, exploring, and comparing explainers for transformer-based language models.
Future work will focus on developing Virgil into a central community resource that organizes explainers for transformer-based language models, contributes to the democratization of explainability, and facilitates the translation of research advances into practice.

\begin{credits}
\subsubsection{\ackname} 
This work was partially supported by the ERC Advanced Grant No. 834756 (“XAI – Science and Technology for the Explanation of AI Decision Making”), by the European Union under Grant Agreement No. 101120763 (“TANGO”), and by FAIR – Future Artificial Intelligence Research (PE00000013, Spoke 1 “Human-Centered AI”), funded under the Next Generation EU programme.

\subsubsection{\discintname}
The authors have no competing interests to declare that are relevant to
the content of this article.
\end{credits}

\bibliographystyle{splncs04}
\bibliography{references}
\end{document}